\pdfoutput=1
\documentclass[11pt]{article}

\usepackage[final]{acl}

\usepackage{times}
\usepackage{latexsym}
\usepackage{booktabs}
\usepackage[T1]{fontenc}
\usepackage[utf8]{inputenc}

\usepackage{microtype}

\usepackage{inconsolata}

\usepackage{graphicx}

\usepackage{url}
\usepackage{tikz}
\usepackage{tabularx}

\usepackage{adjustbox}
\usepackage{xcolor}
\usepackage{hyperref}
 \definecolor{darkblue}{rgb}{0, 0, 0.5}
  \hypersetup{colorlinks=true, citecolor=darkblue, linkcolor=darkblue, urlcolor=darkblue}

\usepackage{xstring}

\usepackage{color}

\title{LLMsAgainstHate @ NLU of Devanagari Script Languages 2025: Hate Speech Detection and Target Identification in Devanagari Languages via Parameter Efficient Fine-Tuning of LLMs}

\author{
 \textbf{Rushendra Sidibomma\textsuperscript{1}},
 \textbf{Pransh Patwa\textsuperscript{2}},
 \textbf{Parth Patwa\textsuperscript{3}},
\\
 \textbf{Aman Chadha\textsuperscript{4,5*}},
  \textbf{Vinija Jain\textsuperscript{4}},
 \textbf{Amitava Das\textsuperscript{6}},
\\
\\
 \textsuperscript{1}IIIT Sri City, India,
 \textsuperscript{2}Aditya English Medium School, India,
 \textsuperscript{3}UCLA, USA, \\
 \textsuperscript{4}Stanford University, USA,
 \textsuperscript{5}Amazon GenAI, USA,
 \textsuperscript{6}University of South Carolina, USA
\\ 
rushendra.s20@iiits.in, 
pransh.patwa@aemspune.edu.in, 
parthpatwa@g.ucla.edu\\
hi@aman.ai, hi@vinija.ai, 
amitava@mailbox.sc.edu
\\
}

\begin{document}
\maketitle
\renewcommand{\thefootnote}{\fnsymbol{footnote}}
\footnotetext[1]{Work does not relate to position at Amazon.}
\renewcommand*{\thefootnote}{\arabic{footnote}}
\setcounter{footnote}{0}
\begin{abstract}
The detection of hate speech has become increasingly important in combating online hostility and its real-world consequences. Despite recent advancements, there is limited research addressing hate speech detection in Devanagari-scripted languages, where resources and tools are scarce. While large language models (LLMs) have shown promise in language-related tasks, traditional fine-tuning approaches are often infeasible given the size of the models. In this paper, we propose a Parameter Efficient Fine tuning (PEFT) based solution for hate speech detection and target identification. We evaluate multiple LLMs on the Devanagari dataset provided by \cite{thapa2025nludevanagari}, which contains annotated instances in 2 languages - Hindi and Nepali. The results demonstrate the efficacy of our approach in handling Devanagari-scripted content. Our code is available at \href{https://github.com/Rushendra10/Hate-Speech-Detection-and-Target-Identification-in-Devanagari-Languages}{https://github.com/Rushendra10/Hate-Speech-Detection-and-Target-Identification-in-Devanagari-Languages}.
\end{abstract}
\section{Introduction}

In recent years, the rise in online hate speech has led to severe social consequences, often escalating into real-world violence and disproportionately affecting vulnerable communities \cite{laub2019hate}. This issue is especially challenging for low-resource languages, where the lack of technological tools limits effective monitoring and mitigation of harmful content \cite{shen2024languagebarrierdissectingsafety,court2024shortcomingsllmslowresourcetranslation}. Addressing hate speech in these languages is important to minimize societal harm and foster safer online environments.

Large Language Models (LLMs) have shown significant potential in handling various language-related tasks, including hate speech detection. However, techniques such as in-context learning (ICL) are increase the cost and latency of LLMs with the increase in data \cite{NEURIPS2022_0cde695b}. While fine-tuning can improve performance, it remains resource-intensive, given the billions of parameters of LLMs. To address these challenges, Parameter-Efficient Fine-Tuning (PEFT) has emerged as a more adaptable and cost-effective solution, making it a compelling choice for this application \cite{patwa2024enhancinglow}.

In this paper, we present our system for detection hate-speech in Devanagari-scripted languages. Our key contributions are: 
\begin{itemize}
    \item We introduce a PEFT-based system for detecting hate speech and identifying targeted individuals or groups.
    \item We evaluate the effectiveness of various LLMs in this context.
    \item We focus on Devanagari-scripted languages, but our system can be potentially applied to other languages as well. 
\end{itemize}

\section{Related Work}
\label{sec:relatedwork}

Detecting hate speech online has become a critical issue due to the potential for real-world consequences. Traditional research in this area focused primarily on high-resource languages like English, where robust datasets and NLP tools facilitated effective models \cite{davidson2017automated,fortuna2018survey}. However, applying these methods to low-resource languages remains a significant challenge due to limited annotated datasets and language-specific resources. For instance, recent research on hate speech detection in Hindi, a low-resource language despite its global prevalence, has highlighted the importance of building dedicated datasets and methodologies tailored to these linguistic contexts \cite{hhsd}. 

Efforts to address these challenges have led to new datasets such as IEHate \cite{jafri2023uncovering}, which specifically captures hate speech in the political discourse of the Indian Assembly Elections. This dataset provides valuable insights and benchmarks for hate speech in low-resource languages, underscoring the need for refined algorithms and hybrid human-machine approaches. Similarly, the HHSD \cite{hhsd} dataset offers a multi-layer annotated dataset for hate speech detection in Hindi, structured hierarchically to categorize hate speech into explicit and implicit forms and target attributes. This dataset demonstrates how multi-task learning (MTL) frameworks, which combine similar tasks across related languages, can improve performance, further advancing hate speech detection in resource-limited languages.

Researchers have attempted hate-speech detection in low resources languages using various deep learning techniques. Some of the languages explored include Bengali \cite{safi-samghabadi-etal-2020-aggression,das-etal-2022-hate-speech}, Hindi \cite{patwa2021hater,10037649,velankar2021hateoffensivespeechdetection,patwa2021overview}, Dravidian languages \cite{tula2021bitions,10419328,tula2022offence} etc. Some researchers have also explored multi-modal low resource hate-speech detection \cite{mishra2023memotion,mishra2023overview,guo2023nuaaqmulaiitmemotion3multimodal}. For a detailed discussion, please refer to \cite{parihar2021hate}.

Large Language Models (LLMs) have improved detection capabilities but require considerable resources for fine-tuning. Parameter-Efficient Fine-Tuning (PEFT) techniques allow for efficient adaptation by tuning only a subset of model parameters, making them practical for low-resource settings \cite{li2021prefixtuningoptimizingcontinuousprompts,lester-etal-2021-power}. Language-agnostic models, leveraging machine translation to standardize inputs, also show promise in multilingual hate speech detection \cite{10.1145/3409334.3452077}.

In-context learning (ICL) has been explored for adapting LLMs without full retraining, though it incurs higher inference costs as examples scale \cite{brown2020languagemodelsfewshotlearners}. In contrast, PEFT methods offer scalable adaptation \cite{liu2022fewshot,patwa2024enhancinglow}, supporting efficient hate speech detection across languages with fewer resources. In our work, we explore LoRA \cite{hu2021lora} for hate speech detection in Devanagari languages Hindi and Nepali.

\section{Data}
\label{sec:data}
We use the dataset released as a part of the shared task \cite{thapa2025nludevanagari} in the CHiPSAL workshop \cite{sarves2025chipsal}. It contains two tasks - hate speech detection and hate speech target identification in two Devanagari scripted languages: Hindi \cite{jafri2024chunav,jafri2023uncovering} and Nepali \cite{thapa2023nehate,rauniyar2023multi}.

\subsection{Hate Speech Detection}
For hate speech detection, the data consists of devanagari-scripted text annotated into 2 classes - hate speech and not hate speech. The texts are diverse and collected from various sources  including social media posts, news articles, and forums, reflecting a wide range of topics and styles. Table \ref{tab:dd_hate} shows the data distribution. We can see that there is a significant class imbalance towards the non-hate class. This imbalance poses a challenge for training the models, as they may tend to favor the majority class.

\begin{table}[h!]
\centering
\begin{tabular}{lrrr}
\toprule
\textbf{Class} & \textbf{Train} & \textbf{Valid} & \textbf{Test} \\
\midrule
Not Hate & 16805 & 3602 & 3601 \\
Hate & 2214 & 474 & 475\\
\midrule
\textbf{Total} & \textbf{19019} & \textbf{4076} & \textbf{4076} \\
\bottomrule
\end{tabular}
\caption{Data distribution of the hate speech detection dataset. }
\label{tab:dd_hate}
\end{table}

\subsection{Hate Speech Target Identification}
The second subtask focuses on identifying the targets of hate speech in Devanagari-scripted text. The goal is to classify whether hate speech is directed towards an individual, an organization, or a community. The dataset for this task contains text samples annotated with target labels. The distribution of targets, as indicated in Table \ref{tab:dd_target}, shows a more balanced representation for individual and organizational targets, with approximately equal numbers of samples for both classes. However, there is a notable scarcity of samples where the target is a community, resulting in a skewed distribution towards individual and organizational targets. This data limitation introduces a potential challenge in accurately predicting hate speech directed at communities.

\begin{table}[h!]
\centering
\begin{tabular}{lrrr}
\toprule
\textbf{Class} & \textbf{Train} & \textbf{Valid} & \textbf{Test} \\
\midrule
Individual & 1074 & 230 & 230 \\
Organizational & 856 & 183 & 184 \\
Community & 284 & 61 & 61\\
\midrule
\textbf{Total} & \textbf{2214} & \textbf{474} & \textbf{475} \\
\bottomrule
\end{tabular}
\caption{Data distribution of the hate speech target identification dataset.}
\label{tab:dd_target}
\end{table}

\section{Methodology}
LLMs leverage the transformer \cite{vaswani2023attentionneed} architecture to model linguistic patterns across vast corpora, utilizing multi-head self-attention mechanisms to capture both local and global dependencies in text. LLMs have billions of parameters and are pretrained on extensive general-purpose corpora. As a result they demonstrate great zero shot capabilities on many natural language tasks \cite{kojima2022large}. However, they struggle on low resource languages \cite{cassano2024knowledge}. 

ICL is a way to improve performance of LLMs. It refers to providing few labeled examples in the prompt to guide the LLM. However, as the number of examples increase, the cost and lantency of inference increases \cite{NEURIPS2022_0cde695b}.

Fully fine tuning (FFT) an LLM with billions of parameters is infeasible because of the costs and computational resources needed \cite{xu2023parameterefficient}. 

Parameter Efficient Fine Tuning (PEFT) is a method in which we only finetune a small number of parameters as compared to the size of the LLM. It is more effective than ICL while being more efficient than FFT \cite{xu2023parameterefficient}.  

For our system we use a PEFT method called Low Rank Adaptation (LoRA) \cite{hu2021lora}.  LoRA reduces the number of trainable parameters by decomposing weight updates into low-rank matrices, which are inserted into the model's attention layers. Specifically, for a weight matrix $W$, LoRA approximates the update as:

\begin{equation}
W' = W + \Delta W = W + A B^T
\end{equation}

where A and B are low-rank matrices. By freezing the core parameters of the pretrained model and only updating the low-rank matrices during training, LoRA significantly decreases computational and memory requirement for training while being as effective as FFT \cite{hu2021lora}. Furthermore, LoRA does not add to the inference latency, as after training, the weight update $AB^T$ is added to the model weights, hence the total number of model weights remains the same.

\section{Experiments}
\label{sec:experiments}
We conduct experiments on 4 different LLMs to address challenges in processing Devanagari-scripted languages. The considered models include the LLama-3.1-8B \cite{dubey2024llama3herdmodels}, Nemo-Instruct-2407 \cite{mistral_nemo_instruct_2407}, Qwen2.5-7B-Instruct \cite{yang2024qwen2technicalreport}, and Phi3-medium-4k-Instruct \cite{abdin2024phi3technicalreporthighly}.  Each model is fine-tuned using task-specific datasets. Quantization of the models to 4-bit precision was employed to reduce memory consumption and to speed up training and  inference. All fine-tuning models used LoRA with $rank = 16$, $alpha = 16$ and no dropout. 

All fine-tuning experiments are performed using a 16GB NVIDIA T4 GPU. For the hate speech detection task, all models were fine-tuned for 2 epochs. For the target identification task, models were fine-tuned for 4 epochs in order to accommodate a relatively small training set. The code is implemented using the Unsloth \cite{unsloth} library, which helps accelerate training.
Our code is available at \href{https://github.com/Rushendra10/Hate-Speech-Detection-and-Target-Identification-in-Devanagari-Languages}{https://github.com/Rushendra10/Hate-Speech-Detection-and-Target-Identification-in-Devanagari-Languages}.

\section{Results and Analysis}
\label{sec:results}

The test performance of the models for the hate speech detection and target identification tasks are provided in Tables \ref{tab: task_b_acc} and \ref{tab: task_c_acc} respectively. We can see that for both the tasks Nemo has the best performance (F1 scores 90.05\% and 71.47\% respectively). Notably, Nemo performs better than Llama despite having smaller size. Furthermore, we can see that the overall performance is better on hate speech detection as compared to target identification. This is because the latter task has 3 classes whereas the former task has only 2 classes.

\begin{table}[]
\centering
\begin{tabular}{cccc}
%\hline
\toprule
\textbf{Model}    & \multicolumn{1}{c}{\textbf{ Size}} & \textbf{Acc.} & \textbf{F1} \\ %\hline
\midrule
Llama-3.1          & 8.03B                                       & 88.71\%             & 88.02\%    \\ %\hline
Phi-3-medium   & 7.36B                                       & 90.06\%             & 88.91\%    \\ %\hline
Qwen2.5        & 4.46B                                       & 88.62\%             & 87.90\%    \\ %\hline
Nemo & 6.97B                                       & \textbf{90.75\%}             & \textbf{90.05\%}    \\ %\hline
\bottomrule
\end{tabular}
\caption{Performance of various models for hate speech detection task on the test set, along with the quantized model size. Acc. refers to accuracy. F1 refers to weighted average F1 score.}
\label{tab: task_b_acc}
\end{table}

\begin{table}[]
\centering
\begin{tabular}{cccc}
%\hline
\toprule
\textbf{Model}    & \multicolumn{1}{c}{\textbf{Size}} & \textbf{Acc.} & \textbf{F1} \\ %\hline
\midrule
Lama-3.1         & 8.03B                                       & 67.37\%             & 66.58\%    \\ %\hline
Phi-3-medium   & 7.36B                                       & 68.21\%             & 67.80\%    \\ %\hline
Qwen2.5        & 4.46B                                       &70.32\%             & 70.41\%    \\\ %\hline
Nemo & 6.97B                                       & \textbf{72.00\%}             & \textbf{71.47\%}    \\ %\hline
\bottomrule
\end{tabular}
\caption{Performance of various models for  target identification task on the test set along with the quantized model size. Acc. refers to accuracy. F1 refers to weighted average F1 score.}
\label{tab: task_c_acc}
\end{table}

\begin{figure}[h]
    \centering\
    \includegraphics[width=\columnwidth]{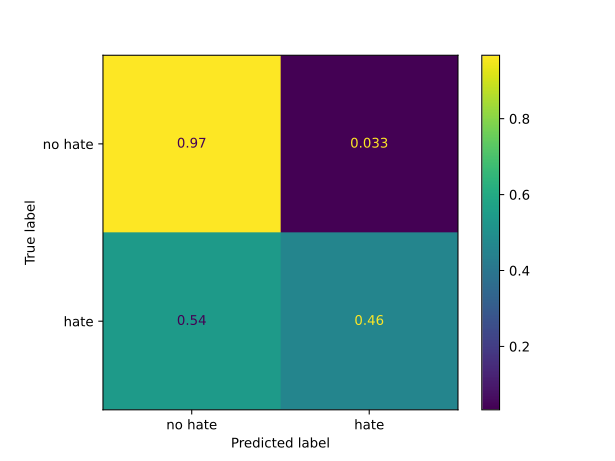}
    \caption{Confusion matrix of Nemo on the test set for hate speech detection.}   
    \label{fig:task_b_conf_matrix}
\end{figure}

\begin{figure}[h]
    \centering
    \includegraphics[width=\columnwidth]{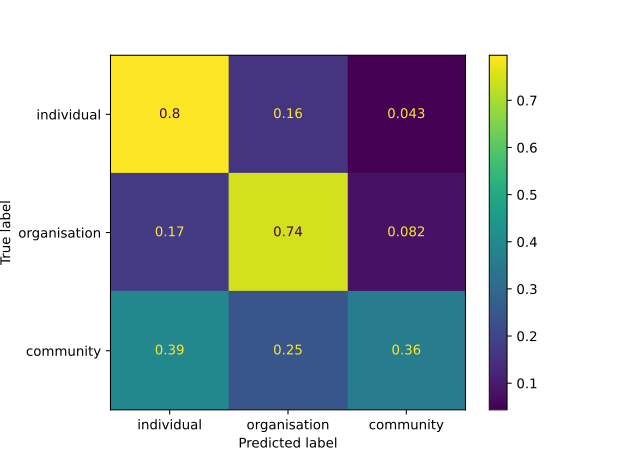}
    \caption{Confusion matrix of Nemo on the test set for hate speech target identification.}
    \label{fig:task_c_conf_matrix}
\end{figure}

\subsection{Class-wise Analysis}

Table \ref{tab: task_b_cls_report} shows the class-wise results of Nemo for hate speech detection task. The F1 score on the hate class is much lower than on the non-hate class. The Confusion Matrix (Figure \ref{fig:task_b_conf_matrix}) shows that the instances of hate class are often mis-predicted as Non Hate. These observations can be attributed to the class imbalance in the training dataset.  

Table \ref{tab: task_c_cls_report} shows the class-wise results of Nemo for hate target identification task. The F1 on Individual class is comparable to that in Organization class, whereas it is significantly lower for the Community class.  From the Confusion Matrix (Figure \ref{fig:task_c_conf_matrix}), we can see that instances of hate directed towards community are frequenty mis-predicted into one of the other 2 classes. Similar to the hate speech detection task, these observation are also a result of the imbalanced training dataset.

\begin{table}[ht!]
\centering
\begin{tabular}{lccc}
\toprule
 & \textbf{P} & \textbf{R} & \textbf{F1} \\
\midrule
\textbf{Non Hate} & 93.10\% & 96.70\% & 94.86\% \\
\textbf{Hate} & 64.58\% & 45.68\% & 53.51\%  \\

\bottomrule
\end{tabular}
\caption{Class-wise performance of Nemo on test set of the hate speech detection task. P = Precision, R= Recall, F1 = F1 score.}
\label{tab: task_b_cls_report}
\end{table}

\begin{table}[ht!]
\centering
\resizebox{\columnwidth}{!}{%
\begin{tabular}{lccc}
\toprule
 & \textbf{P} & \textbf{R} & \textbf{F1}  \\
\midrule
\textbf{Individual} & 76.57\% & 79.56\% & 78.04\%\\
\textbf{Organization} & 72.49\% & 74.46\% & 73.46\%\\
\textbf{Community} & 46.81\% & 36.07\% & 40.74\% \\
\bottomrule
\end{tabular}%
}
\caption{Class-wise performance of Nemo on test set of the target identification task. P = Precision, R= Recall, F1 = F1 score. }
\label{tab: task_c_cls_report}
\end{table}

\section{Conclusion and Future Work}
\label{sec:conclusion}
In this study, we present our approach for hate speech detection in Devanagari-scripted languages using LLMs fine-tuned with LoRA. Our methodology demonstrates good performance, as evidenced by accuracy and F1 score metrics. By leveraging the CHiPSal dataset, we effectively address the challenges posed by low-resource languages. We notice that the performance is lower on the the classes with fewer data instances.

Future research could involved enhancing the model's capabilities by developing data generation techniques to address class imbalance, ensuring robust performance across all classes.
Additionally, investigating the integration of more sophisticated techniques, such as ensemble methods, can further boost detection accuracy and robustness.

\section{Limitation}
We assume the existence of a decently sized train dataset to fine-tune our model. Further, we assume that the LLMs will have some knowledge of devanagari languages for PEFT to work. 

\section{Ethical Statement}
Hate speech detection is a sensitive topic and can be subjective. LLMs are known to have inherent biases. Any censoring decisions based on the LLMs predictions should involve comprehensive human reviews.

\bibliography{acl_latex}
\end{document}